\documentclass[
]{ceurart}

\usepackage{wasysym}
\usepackage{lingmacros}

\begin{document}

\copyrightyear{2020}
\copyrightclause{Copyright for this paper by its authors.\\
  Use permitted under Creative Commons License Attribution 4.0
  International (CC BY 4.0).}

\conference{FIRE 2020: Forum for Information Retrieval Evaluation,
  December 16-20 December, 2020, Hyderabad, India}

\title{WLV-RIT at HASOC-Dravidian-CodeMix-FIRE2020: Offensive Language Identification in Code-switched YouTube Comments}

\author[1]{Tharindu Ranasinghe}
\address[1]{University of Wolverhampton, UK}
\ead{T.D.RanasingheHettiarachchige@wlv.ac.uk}

\author[2]{Sarthak Gupte}
\ead{sg7179@rit.edu}
\author[2]{Marcos Zampieri}
\ead{marcos.zampieri@rit.edu}
\author[2]{Ifeoma Nwogu}
\ead{ion@cs.rit.edu}
\address[2]{Rochester Institute of Technology, USA}

% \ead{marcos.zampieri@rit.edu}
% \ead{sg7179@rit.edu}
% \ead{ion@cs.rit.edu}

% \author[2]{Sarthak Gupte}
% \address[2]{Rochester Institute of Technology, USA}
% \ead{sg7179@rit.edu}

% \author[4]{Ifeoma Nwogu}
% \address[2]{Rochester Institute of Technology, USA}
% \ead{ion@cs.rit.edu}

\begin{abstract}
 This paper describes the WLV-RIT entry to the Hate Speech and Offensive Content Identification in Indo-European Languages (HASOC) shared task 2020. The HASOC 2020 organizers provided participants with annotated datasets containing social media posts of code-mixed in Dravidian languages (Malayalam-English and Tamil-English). We participated in task 1: Offensive comment identification in Code-mixed Malayalam Youtube comments. In our methodology, we take advantage of available English data by applying cross-lingual contextual word embeddings and transfer learning to make predictions to Malayalam data. We further improve the results using various fine tuning strategies. Our system achieved 0.89 weighted average F1 score for the test set and it ranked 5$^{th}$ place out of 12 participants. 
\end{abstract}


\maketitle

\section{Introduction}

Offensive content is pervasive in social media putting users of various platforms at risk \cite{bauman2013associations}. The pervasiveness of such content motivated the development of several systems capable of identifying offensive posts in a number of languages \cite{kumar2020evaluating,pitenis2020}. Once identified, these posts can be then set aside for human moderation or deleted from online platforms mitigating risks to their users \cite{risch-krestel-2018-delete}. 

Recent studies have addressed many types of offensive content such as online abuse \cite{nobata2016abusive,founta2018large}, aggression \cite{kumar2018benchmarking}, cyberbullying \cite{chelmis2017mining,yao2018cyberbullying}, and hate speech \cite{malmasi2017detecting,mathew2019temporal}. International workshops and competitions such as HatEval 2019 \cite{basile2019semeval}, OffensEval 2019 and 2020 \cite{offenseval,offenseval2020}, co-located with SemEval, have been organized in the last two years attracting a large number of participants. Most high performing system in these competitions used neural networks and contextual word embeddings such as BERT \cite{devlin2019bert}.

In this paper we describe the WLV-RIT entry to the the HASOC 2020 shared task which featured Malayalam-English code-switched data. Building on the experience of recent high performing models submitted to the OffensEval competitions, we use a transformer-based architecture described in detail in Section \ref{sec:method}. The HASOC 2020 code-switching dataset is a particularly challenging one for state-of-the-art offensive language detection systems and we make use of transfer learning techniques that have been recently applied to project predictions from English to resource-poorer languages with great success  \cite{ranasinghe-etal-2020-multilingual}.

%\section{Related Work}

\section{Task description and Datasets}
\label{sec:description}

The goal of this task is to identify the offensive language of the code-mixed dateset of comments/posts in Dravidian Languages (Tamil-English and Malayalam-English) collected from social media. Each comment/post is annotated with an offensive language label at the comment/post level. The dataset has been collected from YouTube comments \cite{hasocdravidian-ceur}. We participated in task 1 which is a message-level label classification task; given a YouTube comment in Code-mixed (Mixture of Native and Roman Script) Tamil and Malayalam, systems have to classify whether a post is offensive or not-offensive. To the best of our knowledge, this is the first dataset to be released for offensive language detection in Dravidian Code-Mixed text \cite{hasocdravidian-ceur}.

In addition to the dataset provided by the organisers we also used an English Offensive Language Identification Dataset (OLID) \cite{zampieri2019predicting} used in the SemEval-2019 Task 6 (OffensEval) \cite{offenseval} for transfer learning experiments which are describing in Section \ref{sec:method}. OLID is arguably one of the most popular offensive language datasets. It contains manually annotated tweets with the following three-level taxonomy and labels:

\begin{itemize}
    \item[A:] Offensive language identification - offensive vs. non-offensive;
    \item[B:] Categorization of offensive language - targeted insult or thread vs. untargeted profanity;
    \item[C:] Offensive language target identification - individual vs. group vs. other.
\end{itemize}

\noindent We adopted the transfer learning strategy similar to previous recent work \cite{ranasinghe-etal-2020-multilingual}. We believe that the flexibility provided by the hierarchical annotation model of OLID allows us to map OLID level A (offensive vs. non-offensive ) to labels in the HASOC Malayalam-English dataset. 

\section{Methods}
\label{sec:method}

The methodology applied in this work is divided in two parts. Subsection \ref{subsec:traditional} describes traditional machine learning applied to this task and in Subsection \ref{subsec:classification} we describe the transformer models used.

The motivation behind our methodology is the recent success that the transformers had in wide range of NLP tasks like language generation  \cite{devlin2019bert}, sequence classification \cite{ranasinghe-etal-2020-brums,ranasinghe2019brums}, word similarity \cite{hettiarachchi-etal-2020-brums}, named entity recognition \cite{10.1145/3394486.3403149}, question and answering \cite{yang-etal-2019-end} etc. The main idea of the methodology is that we train a classification model with several transformer models in-order to identify offensive texts. However, the transformer models are known to be resource intensive requiring fairly large datasets \cite{devlin2019bert}, therefore, we also experimented with several traditional machine learning models

\subsection{Traditional Machine Learning Methods}
\label{subsec:traditional}
%Include traditional ML methods (Random Forest, SVM, Naive Bayes).
In the first part of the methodology, we used traditional machine learning models. We experimented with three models; Multinomial Naive Bayes \cite{kibriya2004multinomial}, Support Vector Machines \cite{cortes1995support}, and Random Forest \cite{liaw2002classification}. The models take an input vector created using Bag-of-words and outputs a label, either offensive or non-offensive. The models for Multinomial Naive Bayes, SVM and Random Forest were implemented using the Scikit-learn \cite{pedregosa2011scikit}. %\footnote{Scikit Learn Documentation - \url{https://scikit-learn.org/stable/}} 

% The task defined in the problem statement is classification task and each model works best for binary classification.

\vspace{-2mm}

\paragraph{Data Preprocessing} We performed three preprocessing techniques; removing punctuations, removing emojis and lemmatising the English words. This was done with the use of the NLTK (Natural Language Toolkit) library \cite{bird2009natural} in Python.
%\footnote{NLTK Library - \url{https://www.nltk.org/}}

% Data preprocessing was kept minimal as the dataset is in multilingual English and Malayalam. Performing multiple preprocessing technique will lead to loss of information because the tweets were written mostly in Malayalam language. The preprocessing performed was focused on removal of punctuation's, emojis, converting English words to their root word also known as lemmatization. This was done with the use of nltk(Natural Language Toolkit) library \footnote{NLTK Library - \url{https://www.nltk.org/}} in Python language. Since the traditional ML models works better with the number we used Bag-of-words model to convert the tweets into the vector which was further provided as input to the models. The bag-of-words model create a vocabulary vector containing all the unique words in the dataset and the tweets are replaced by the vector where the words present in the tweets are replaced by 1 and the words in the vector which are not present in the tweet are replaced by 0.

\vspace{-2mm}

\paragraph{Hyper Parameter Optimisation}
% Hyper parameters are the parameters used by the model during the training or learning process to improve the output of the model and better training of the model. 
Optimisation of hyper parameters was performed on SVM and random forest only. For SVM, the hyper parameters fine-tuned were alpha, random state and max iteration, where alpha represents regularisation, random state is used for shuffling of the data and max iteration denotes number of passes through the training data which is also known as epochs. Optimal values achieved were alpha=0.001, random state=5, max iteration=15. For random forest, only one hyper parameter was used which is n-estimator that denotes number of decision trees created. Optimal value achieved for number of trees was 500.

% Label encoders was used on the label column or target column to for prediction where 0 refers to not-offensive and 1 refers to offensive, this process is also called as one hot encoding. The models were fairly quick to predict the output of the given input tweets.

\subsection{Transformers Models}
\label{subsec:classification}
\begin{figure}[pos=!ht]
\centering
\includegraphics[scale=0.38]{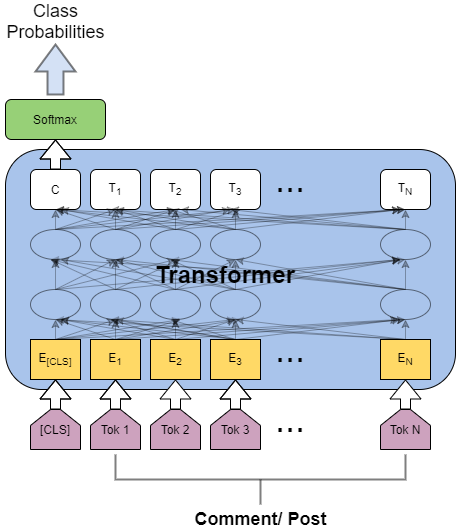}
\caption{Transformer Text Classification Architecture}
\label{fig:architecture}
\end{figure}

As the second part of the methodology, we used Transformer models. Transformer architectures have been trained on general tasks like language modelling and then can be fine-tuned for classification tasks \cite{10.1007/978-3-030-32381-3_16}. They take an input of a sequence and outputs the representation of the sequence. The sequence has one or two segments that the first token of the sequence is always [CLS] which contains the special classification embedding and another special token [SEP] is used for separating segments. For text classification tasks, Transformer models take the final hidden state \textbf{h} of the first token [CLS] as the representation of the whole sequence \cite{10.1007/978-3-030-32381-3_16}. A simple softmax classifier is added to the top of the transformer model to predict the probability of a class as shown in Equation \ref{equ:softmax} where W is the task-specific parameter matrix. The architecture diagram of the classification is shown in Figure \ref{fig:architecture}

\begin{equation}
\label{equ:softmax}
p(c|\textbf{h}) = softmax(W\textbf{h}) 
\end{equation}

\paragraph{Transformers}
\label{subsec:transformers}
We experimented two pretrained transformer models; BERT \cite{devlin2019bert} and XLM-ROBERTA \cite{conneau2019unsupervised} We used the HuggingFace's implementation of the transformer models \cite{Wolf2019HuggingFacesTS} and the pre-trained models available in the HuggingFace model repository.\footnote{HuggingFace model repository - \url{https://huggingface.co/models}} These models were used mainly considering their support to Malayalam language. For BERT we used the BERT multilingual model (BERT-M) and for XLM-ROBERTA (XLM-R) we used the XLM-R-Large model. Both models support 104 languages including Malayalam.  The interesting fact about XLM-R is that it is very compatible in monolingual benchmarks while achieving best results in cross-lingual benchmarks at the same time \cite{conneau2019unsupervised}.

\vspace{-2mm}

\paragraph{Transfer Learning}
%\label{subsec:transfer}
The main idea of the transfer learning strategy is that we train a classification model on a resource rich language, typically English, using a transformer model and perform transfer learning on a less resource language. We trained the classification model on the first level of OLID \cite{OLID} and then we save the weights of the transformer model as well as the softmax layer. We use this saved weights from English to initialise the weights when we are training the classification model for Malayalam. This strategy has improved the performance of different languages with less resources for offensive language identification such as Hindi, Bengali etc \cite{ranasinghe-etal-2020-multilingual}. Therefore we experimented with this strategy to see whether it improves the results for Malayalam too. According to the recent research, cross-lingual transformers have slight edge when using this transfer-learning strategy \cite{ranasinghe-etal-2020-multilingual}. 

\vspace{-2mm}

\paragraph{Data Preprocessing}
\label{subsec:data_preprocessing}
The data preprocessing for this task was kept fairly minimal to make it portable for other languages too. We only followed one data preprocessing technique; converting emojis to text.
Emojis are found to play a key role in expressing emotions in the context of social media \cite{hettiarachchi2019emoji}. But, we cannot assure the existence of embeddings for emojis in pretrained models. Therefore as a preprocessing step, we converted emojis to text. For this conversion we used the Python libraries \textit{demoji} \footnote{demoji repository - \url{https://github.com/bsolomon1124/demojis}} and \textit{emoji} \footnote{emoji repository - \url{https://github.com/carpedm20/emoji}}. \textit{demoji} returns a normal descriptive text and \textit{emoji} returns a specifically formatted text. For an example, the conversion of \smiley \space is `slightly smiling face' using \textit{demoji} and `:slightly\textunderscore smiling\textunderscore face:' using \textit{emoji}. Considering that \textit{demoji} returns a normal text, we used \textit{demoji} to convert the emojis to text.

\vspace{-2mm}

\paragraph{Fine-tuning}
\label{subsec:fine_tuning}
To improve the models, we experimented different fine-tuning strategies: majority class self-ensemble, average self-ensemble, language modelling, which are described below. These fine tuning strategies have shown promising results in recent shared tasks \cite{hettiarachchi-2020-informiner}.  

\begin{enumerate}
\item{\textbf{Self-Ensemble (SE)}} - Self-ensemble is found as a technique which results better performance than the performance of a single model \cite{xu2020improving}. In this approach, same model architecture is trained or fine-tuned with different random seeds or train-validation splits. Then the output of each model is aggregated to generate the final results. As the aggregation methods, we analysed majority-class and average in this research. The number of models used with self-ensemble will be denoted by $N$.
\begin{itemize}
    \item \textit{Majority-class SE (MSE)} - As the majority class, we computed the mode of the classes predicted by each model. Given a data instance, following the softmax layer, a model predicts probabilities for each class and the class with highest probability is taken as the model predicted class. 

    \item \textit{Average SE (ASE)} - In average SE, final probability of class $c$ is calculated as the average of probabilities predicted by each model as in Equation \ref{equ:ase} where \textit{h} is the final hidden state of the [CLS] token. Then the class with highest probability is selected as the final class. 
    
    \begin{equation}
    \label{equ:ase}
    p_{ASE}(c|h) = \frac{\sum_{k=1}^{N} p_{k}(c|h) }{N}
    \end{equation}
\end{itemize}

\item{\textbf{Language Modelling (LM)}} - As language modelling, we retrained the transformer model on task dataset before fine-tuning it for the downstream task; text classification. This training is took place according with the model's initial trained objective. Following this technique model understanding on the task data can be improved. 
\end{enumerate}

\vspace{-2mm}

\paragraph{Implementation}
We used a Nvidia Tesla K80 GPU to train the models. We mainly fine tuned the learning rate and number of epochs of the classification model manually to obtain the best results for the validation set. We obtained $1e^-5$ as the best value for learning rate and 3 as the best value for number of epochs for all the languages. Training for English language took around 1 hour while training for Malayalam took around 30 minutes. 

\section{Results and Evaluation}
\label{sec:results}

In this section, we report the experiments we conducted and their results. As informed by the task organisers, we used Weighted Average F1 score to measure the model performance. We also report Precision, Recall and F1 score for each class label as well the Macro F1 score in the results tables. Results in Tables \ref{table:traditional} - \ref{table:ase_lm} are computed on validation dataset. Finally, in Section \ref{subsec:sub_results} we report the results provided by organisers to our models, for the test set. 

Table \ref{table:traditional} shows the results we gained with traditional machine learning algorithms. Out of the three traditional machine learning algorithms Random Forest performed best, providing us with 0.93 weighted average F1 score. In the experiments we did with transformers, initially we focused on the impact of transfer learning when used with different transformer models and the obtained results are summarised in Table \ref{table:default}. According to the results XLM-R with transfer learning outperformed other models. Also we could notice that transfer learning improved both models; BERT and XLM-R. 

\begin{table*}[pos=!ht]
%\footnotesize
\centering
\scalebox{0.95}{
\begin{tabular}{l|ccc|ccc|ccc|c}

\hline
                                     & \multicolumn{3}{c|}{\textbf{Non Hate Offensive}} & \multicolumn{3}{c|}{\textbf{Hate Offensive}}             & \multicolumn{3}{c|}{\textbf{Weighted Average}}      & \textbf{}         \\ \hline
\multicolumn{1}{l|}{\textbf{Model}} & \textbf{P}   & \textbf{R}   & \textbf{F1}   & \textbf{P} & \textbf{R} & \textbf{F1}               & \textbf{P} & \textbf{R} & \textbf{F1}               & \textbf{F1 Macro} \\ \hline
\textit{Random Forest}                      & 0.93         & 0.99         & 0.96          & 0.92       & 0.68       & 0.78 & 0.93       & 0.93       & \textbf{0.93}  & 0.87   \\

\textit{Linear SVM}                           & 0.93         & 0.98         & 0.96          & 0.88       & 0.68       & 0.77 & 0.92       & 0.93       & 0.92           & 0.86   \\
\textit{Mult. Naive Bayes}                     & 0.90         & 0.98         & 0.94          & 0.88       & 0.53       & 0.66 & 0.90       & 0.90       & {0.89}           & 0.80   \\

\hline
\end{tabular}
}
\caption[Results for traditional ML models]{Results for offensive language detection with traditional ML models. For each model, Precision (P), Recall (R), and F1 are reported on all classes, and weighted averages. Macro-F1 is also listed.}
\label{table:traditional}
\end{table*}

% \subsection{Impact by Transfer Learning}
% \label{subsec:impact_transfer}

\begin{table*}[pos=!ht]
%\footnotesize
\centering
\scalebox{0.95}{
\begin{tabular}{l|ccc|ccc|ccc|c}

\hline
                                     & \multicolumn{3}{c|}{\textbf{Non Hate Offensive}} & \multicolumn{3}{c|}{\textbf{Hate Offensive}}             & \multicolumn{3}{c|}{\textbf{Weighted Average}}      & \textbf{}         \\ \hline
\multicolumn{1}{l|}{\textbf{Model}} & \textbf{P}   & \textbf{R}   & \textbf{F1}   & \textbf{P} & \textbf{R} & \textbf{F1}               & \textbf{P} & \textbf{R} & \textbf{F1}               & \textbf{F1 Macro} \\ \hline
\textit{XLM-R (TL)}                      & 0.91         & 0.96         & 0.94          & 0.77       & 0.59       & 0.67 & 0.89       & 0.89       & \textbf{0.89}  & 0.80   \\
\textit{BERT-m (TL)}                     & 0.90         & 0.95         & 0.93          & 0.76       & 0.57       & 0.65 & 0.88       & 0.88       & 0.87           & 0.78   \\
\textit{XLM-R}                           & 0.89         & 0.98         & 0.93          & 0.79       & 0.40       & 0.53 & 0.88       & 0.88       & 0.86           & 0.74   \\
\textit{BERT-m}                          & 0.88         & 0.97         & 0.92          & 0.78       & 0.38       & 0.51 & 0.86       & 0.87       & 0.85           & 0.72   \\
\hline
\end{tabular}
}
\caption[Results for Default settings with Transformers]{Results for offensive language detection with default settings on Transformers. For each model, Precision (P), Recall (R), and F1 are reported on all classes, and weighted averages. Macro-F1 is also listed. \textit{TL} indicated the \textit{Transfer Learning} experiments}
\label{table:default}
\end{table*}

% \subsection{Impact by Self Ensemble Methods}
% \label{subsec:impact_self_ensemble}
\noindent The self ensemble methods 
%mentioned in Section \ref{subsec:fine_tuning} 
were experimented using all the transformer models and obtained results are summarised in Tables \ref{table:mse} and \ref{table:ase}. In most experiments, ASE has given a higher F1 than MSE and it improved the results over the default settings. With that fine tuning strategy too XLM-R with transfer learning outperformed all the other models.

 The language modeling fine tuning strategy 
%mentioned in Section \ref{subsec:fine_tuning} 
were experimented using all the transformer models and obtained results are summarised in Table \ref{table:ase_lm}. These experimented were done on top of ASE fine tuning strategy since it provided better results than the default settings. Results show that language modeling clearly improved the results. In fact, the best result from our experiments were shown when XLM-R model with transfer learning fine tuned with ASE and language modeling. 

\begin{table*}[pos=htb]
%\footnotesize
\centering
\scalebox{0.95}{
\begin{tabular}{l|ccc|ccc|ccc|c}

\hline
                                     & \multicolumn{3}{c|}{\textbf{Non Hate Offensive}} & \multicolumn{3}{c|}{\textbf{Hate Offensive}}             & \multicolumn{3}{c|}{\textbf{Weighted Average}}      & \textbf{}         \\ \hline
\multicolumn{1}{l|}{\textbf{Model}} & \textbf{P}   & \textbf{R}   & \textbf{F1}   & \textbf{P} & \textbf{R} & \textbf{F1}               & \textbf{P} & \textbf{R} & \textbf{F1}               & \textbf{F1 Macro} \\ \hline
\textit{XLM-R (TL)}                      & 0.91         & 0.96         & 0.94          & 0.77       & 0.59       & 0.67 & 0.89       & 0.89       & \textbf{0.89}  & 0.80   \\
\textit{BERT-m (TL)}                     & 0.90         & 0.95         & 0.93          & 0.76       & 0.57       & 0.65 & 0.88       & 0.88       & 0.87           & 0.78   \\
\textit{XLM-R}                           & 0.89         & 0.98         & 0.93          & 0.79       & 0.42       & 0.55 & 0.88       & 0.88       & 0.86           & 0.76   \\
\textit{BERT-m}                          & 0.88         & 0.97         & 0.92          & 0.78       & 0.40       & 0.53 & 0.87       & 0.87       & 0.85           & 0.74   \\
\hline
\end{tabular}
}
\caption[Results for MSE with Transformers]{Results for offensive language detection with MSE on Transformers. For each model, Precision (P), Recall (R), and F1 are reported on all classes, and weighted averages. Macro-F1 is also listed. \textit{TL} indicated the \textit{Transfer Learning} experiments}
\label{table:mse}
\end{table*}

\begin{table*}[pos=htb]
%\footnotesize
\centering
\scalebox{0.95}{
\begin{tabular}{l|ccc|ccc|ccc|c}

\hline
                                     & \multicolumn{3}{c|}{\textbf{Non Hate Offensive}} & \multicolumn{3}{c|}{\textbf{Hate Offensive}}             & \multicolumn{3}{c|}{\textbf{Weighted Average}}      & \textbf{}         \\ \hline
\multicolumn{1}{l|}{\textbf{Model}} & \textbf{P}   & \textbf{R}   & \textbf{F1}   & \textbf{P} & \textbf{R} & \textbf{F1}               & \textbf{P} & \textbf{R} & \textbf{F1}               & \textbf{F1 Macro} \\ \hline
\textit{XLM-R (TL)}                      & 0.92         & 0.97         & 0.95          & 0.78       & 0.60       & 0.68 & 0.90       & 0.90       & \textbf{0.90}  & 0.81   \\
\textit{BERT-m (TL)}                     & 0.91         & 0.96         & 0.94          & 0.77       & 0.58       & 0.66 & 0.89       & 0.89       & 0.88           & 0.79   \\
\textit{XLM-R}                           & 0.90         & 0.99         & 0.94          & 0.80       & 0.43       & 0.56 & 0.89       & 0.89       & 0.87           & 0.77   \\
\textit{BERT-m}                          & 0.89         & 0.98         & 0.93          & 0.79       & 0.41       & 0.54 & 0.88       & 0.88       & 0.86           & 0.75   \\
\hline
\end{tabular}
}
\caption[Results for ASE with Transformers]{Results for offensive language detection with ASE on Transformers. For each model, Precision (P), Recall (R), and F1 are reported on all classes, and weighted averages. Macro-F1 is also listed. \textit{TL} indicated the \textit{Transfer Learning} experiments}
\label{table:ase}
\end{table*}

\begin{table*}[pos=htb]
%\footnotesize
\centering
\scalebox{0.95}{
\begin{tabular}{l|ccc|ccc|ccc|c}

\hline
                                     & \multicolumn{3}{c|}{\textbf{Non Hate Offensive}} & \multicolumn{3}{c|}{\textbf{Hate Offensive}}             & \multicolumn{3}{c|}{\textbf{Weighted Average}}      & \textbf{}         \\ \hline
\multicolumn{1}{l|}{\textbf{Model}} & \textbf{P}   & \textbf{R}   & \textbf{F1}   & \textbf{P} & \textbf{R} & \textbf{F1}               & \textbf{P} & \textbf{R} & \textbf{F1}               & \textbf{F1 Macro} \\ \hline
\textit{XLM-R (TL)}                      & 0.94         & 0.99         & 0.97          & 0.82       & 0.63       & 0.69 & 0.93       & 0.93       & \textbf{0.93}  & 0.85   \\
\textit{BERT-m (TL)}                     & 0.92         & 0.97         & 0.95          & 0.78       & 0.59       & 0.67 & 0.91       & 0.91       & 0.91           & 0.82   \\
\textit{XLM-R}                           & 0.91         & 0.99         & 0.95          & 0.81       & 0.44       & 0.56 & 0.90       & 0.90       & 0.88           & 0.78   \\
\textit{BERT-m}                          & 0.890         & 0.99         & 0.94          & 0.80       & 0.42       & 0.55 & 0.89       & 0.89       & 0.87           & 0.76   \\
\hline
\end{tabular}
}
\caption[Results for ASE and LM with Transformers]{Results for offensive language detection with ASE and Language Modeling on Transformers. For each model, Precision (P), Recall (R), and F1 are reported on all classes, and weighted averages. Macro-F1 is also listed. \textit{TL} indicated the \textit{Transfer Learning} experiments}
\label{table:ase_lm}
\end{table*}

\subsection{Submission Results}
\label{subsec:sub_results}
Considering the evaluation results on the validation set, we selected the fine-tuned XLM-R(TL) model with ASE + language modeling as our official submission to the HASOC task. According to the results provided by the organisers, our best model has scored 0.89 weighted average F1 score on the test set and ranked 5$^{th}$ out of 12 participants. 

\section{Analysis}
\label{sec:analysis}

In addition to the experiments described in this paper, we carried out a qualitative analysis on the dataset to find interesting patterns and observations. In the training data out of 3,200 tweets only 567 were labelled offensive and the remaining 2,633 were labelled as not-offensive. The use of English words were minimal although there are many tweets which are in Malayalam language but written in Roman script. When analysing the tweets labelled as offensive, we observed that there are many tweets in the dataset which are actually not-offensive but labelled as offensive. Free English translations of some examples include: 
\enumsentence{Spent 4 years proclaiming to be a Royal Mech.}
\vspace{-5mm}
\enumsentence{There are 25k dislikes from Ikka (Mammooty) fans, you are free to unlike and cry.}
\vspace{-5mm}
\enumsentence{Nice, looks like a TV drama series from SuryaTV (a Malayalam channel).}
\vspace{-5mm}
\enumsentence{Have you no shame defaming a reputed hospital?}

\noindent We observed that between 20\% and 25\% of the tweets which are labelled as offensive are similar to the example shown above which has certainly impacted the performance of the models. 

\section{Conclusion}
\label{sec:conc}

In this paper we have presented the system submitted by the WLV-RIT team to the HASOC 2020 - Offensive Language Identification - Dravidian Code Mix Task 1 at FIRE 2020. Following a recent study \cite{ranasinghe-etal-2020-multilingual}, we have shown that the XLM-R with transfer learning is the most successful transformer model from several transformer models we experimented. It should be noted that the traditional machine learning models comes very close to the performance of the transformer models. We have shown that the best traditional machine learning algorithm we experimented; Random Forest outperforms the majority of our transformer model based experiments. This can be due to properties of the dataset or due to the fact that a low-resource language like Malayalam is under represented in multilingual pre-trained models. With several fine tuning strategies, XLM-R with transfer learning provides the best result for the validation set. Finally, our approach achieved 5$^{th}$ place in the leaderboard for the test set.

\section*{Acknowledgements}

We would like to thank the HASOC organizers for running this interesting shared task and for replying promptly to all our inquiries. We further thank the anonymous reviewers for their insightful feedback. 

\bibliography{sample-ceur}

\end{document}